# Towards Robust Stacked Capsule Autoencoder
# with Hybrid Adversarial Training

*Jiazhu Dai, Siwei Xiong*

*School of Computer Engineering and Science, Shanghai University, Shanghai 200444, China*

**Abstract:** Capsule networks (CapsNets) are new neural networks that classify images based on the spatial relationships of features. By analyzing the pose of features and their relative positions, it is more capable to recognize images after affine transformation. The stacked capsule autoencoder (SCAE) is a state-of-the-art CapsNet, and achieved unsupervised classification of CapsNets for the first time. However, the security vulnerabilities and the robustness of the SCAE has rarely been explored. In this paper, we propose an evasion attack against SCAE, where the attacker can generate adversarial perturbations based on reducing the contribution of the object capsules in SCAE related to the original category of the image. The adversarial perturbations are then applied to the original images, and the perturbed images will be misclassified. Furthermore, we propose a defense method called Hybrid Adversarial Training (HAT) against such evasion attacks. HAT makes use of adversarial training and adversarial distillation to achieve better robustness and stability. We evaluate the defense method and the experimental results show that the refined SCAE model can achieve 82.14% classification accuracy under evasion attack. The source code is available at https://github.com/FrostbiteXSW/SCAE_Defense.

## 1. Introduction

Convolutional neural networks (CNNs) perform well when handling tasks of computer vision. Though CNNs have excellent capabilities of fitting and generalization, and can obtain accurate recognition results among various datasets and samples, CNNs are not good at addressing affine transformations such as rotation and resizing.

The original intention of capsule networks (CapsNets) is to solve the problem of CNNs when dealing with affine transformations. CapsNets extract parts of different objects in the image and analyze their relationships in order to reveal the compositions of the objects and their correlations. The stacked capsule autoencoder (SCAE) [1] is a state-of-the-art CapsNet which introduces the theory of CapsNets into autoencoders. By capturing the pose, presence and features of different parts and objects, the SCAE is able to conduct unsupervised classification on images.

However, recent studies have proved that CapsNets are vulnerable to adversarial attacks [2–5], and so is the SCAE. In this paper, we propose an evasion attack against the SCAE. After a perturbation is generated based on the output of the SCAE's object capsules in order to decrease their contribution to correct classification, it is applied to the clean image to form an adversarial one. Then the adversarial image is input into the SCAE and misclassified by the classifiers of the SCAE. According to our studies and experiments, it is possible to construct imperceptible adversarial samples to deceive the SCAE, which may cause its predictions no longer reliable and raise concerns for applying it in safety-critical applications [6].

At the moment, there are studies aiming at improving the resistance of CapsNets to adversarial attacks [7–9], but these studies focus on CapsNets with dynamic routing, and the adversarial robustness of the SCAE has rarely been explored. In this paper, we propose a defense method called Hybrid Adversarial Training (HAT) against the above threat, which is based on adversarial training and adversarial distillation. Adversarial training helps the SCAE to discover and fix the vulnerabilities by inserting adversarial samples into training datasets. Adversarial distillation ensures the classification accuracy of the SCAE on clean samples. HAT combines the above two defense methods, and not only ensure the accuracy, but also further improves the robustness of the SCAE. The experimental results prove that our defense method can enhance the resistance of the SCAE to adversarial attacks, and achieve 82.14% classification accuracy under evasion attack. The contributions of the paper can be summarized as follows:

1. We propose an evasion attack method against the SCAE, which can cause the classifiers of the SCAE to output wrong predictions. The evasion attack method confirms the existence of security vulnerabilities in the SCAE, and it has high attack success rate and stealthiness;

2. We propose a defense method called HAT against the above evasion attack, where neither the original structure of the SCAE is modified, nor new modules are added during the testing phase. Our defense method ensures that the SCAE can maintain relatively high classification accuracy on adversarial samples and meanwhile achieve similar classification accuracy to that of the original model on clean samples;

The remainder of this paper is organized as follows: Section 2 introduces the related works; Section 3 explains the architecture and operations of the SCAE, and the theory of adversarial training and distillation; Section 4 describes the threat model; Section 5 presents our defense method in detail; Section 6 describes the experiments and results of our defense method; and Section 7 provides a summary and briefly presents our future work.

## 2. Related Works

### 2.1 Capsule Networks

CapsNets are a type of models that recognize images according to spatial relationships. So far, CapsNets have developed three different versions: the dynamic routing capsule network proposed by Sabour et al. [10] in 2017, the EM routing capsule network proposed by Hinton et al. [11] in 2018, and the stacked capsule autoencoder proposed by Kosiorek et al. [1] in 2019.

As a state-of-the-art CapsNet, the SCAE is different from the past CapsNets in that it focuses on unsupervised learning and has autoencoders but no routing algorithm in its structure. The SCAE model consists of the part capsule autoencoder (PCAE) and the object capsule autoencoder (OCAE). The PCAE decomposes the input image into small parts, and passes the information of these parts to the OCAE. The OCAE composes the parts into different objects, and output the presence probability of all known objects. Finally, a classifier categorizes the image based on the objects that appear in it.

The PCAE and the OCAE are composed of special neurons called capsules. A part capsule inside the PCAE contains a six-dimensional pose vector, a one-dimensional presence probability and an n-dimensional attribute vector. The object capsules inside the OCAE are encoded by a Set Transformer [12], each of which is composed of multiple part capsules. As for the classifiers, the SCAE uses k-means classifiers for unsupervised classification, and also provides optional linear classifiers for supervised classification.

The main contribution of the SCAE is a new method for image classification, that is, to decompose the image into several small parts and recompose them into bigger objects. During this procedure, the SCAE obtains the relationships between different parts and objects. Unlike CNNs using local features to classify images, the SCAE considers the spatial relationships between features and the variety of representations of similar features so as to suppress the influence of the change of a single feature, which makes the SCAE have better resistance to random perturbations. The SCAE can achieve 98.7% accuracy of unsupervised classification on the MNIST dataset.

## 2.2 Adversarial Attacks

Adversarial attacks against machine learning can be categorized into two types: poisoning attacks and evasion attacks.

Poisoning attacks, which occur during the training phase, aim at degrading the performance of the model or creating backdoors in the model so as to control its behavior. The attacker controls the training process of the model in order to create a backdoor inside it by adding elaborately constructed malicious samples to the training dataset, so as to make the model output the results specified by the attacker on the samples containing the backdoor pattern, or decrease the accuracy of the model at the testing phase [13–22].

Evasion attacks, which occur during the testing phase, aim at creating adversarial samples to deceive the model. The attacker applies an imperceptible perturbation to the clean sample to form an adversarial one. The adversarial sample will be misclassified or categorized as the attacker specifies by the model [23–36]. The defense methods proposed in this paper target on evasion attacks.

## 2.3 Adversarial Attack Defenses

There are two popular ways to defense adversarial attacks: modifying the model, which occurs at the training phase; and filtering adversarial samples, which occurs at the testing phase [37].

Modifying the model improves the resistance of the model to adversarial attacks by adjusting the parameters or the structure of the model. This type of defense methods usually works under white-box condition, which means that the defender needs to know the details of the model. On the other hand, there is no need to add extra modules to the model at the testing phase [7–9,25,26,38–53].

Filtering adversarial samples detects and removes adversarial samples, or processes input samples to remove potential adversarial perturbations. These mechanisms are usually set up in the testing phase and the model remains unmodified, so that they can work under black-box condition, which means that the details of the model are not available to the defender. However, these mechanisms consume extra time, and may affect the performance of the model [54–62].

## 2.4 Security Issues about Capsule Networks

After the emergence of CapsNets, research about their robustness and security application mainly focuses on the dynamic routing capsule network. Frosst et al. [54] detected adversarial samples by measuring the Euclidean distances between input samples and their reconstructions generated by the dynamic routing capsule network. Qin et al. [55] constructed a network with cycle-consistent winning capsule reconstructions based on the dynamic routing capsule network, and used three detectors to filter adversarial samples. Deng et al. [56] filtered adversarial samples using a CapsNet with refined dynamic routing algorithm.

However, CapsNets also face security threats. Jaesik et al. [2] transferred several adversarial attacks to the dynamic routing capsule network and successfully fooled it. Michels et al. [3] proved that the dynamic routing capsule network is not more resistant to white-box adversarial attacks than CNNs. Marchisio et al. [4] designed a black-box adversarial attack against the dynamic routing capsule network and verified its effectiveness on the GTSRB dataset. De Marco [5] proved that the dynamic routing capsule network is vulnerable to adversarial attacks.

As the security of the dynamic routing capsule is questionable, its security applications are no longer reliable. Li et al. [7] improved the robustness of the dynamic routing capsule network by adversarial training. Peer et al. [8] designed a capsule network with $\gamma$-capsules by adding a new inductive bias and replacing the routing algorithm, which is more capable of defending adversarial attacks. Garg et al. [9] enhanced the adversarial robustness of the dynamic routing capsule network via thermometer encoding and adversarial training.

The above studies are limited to CapsNets with routing algorithm. Though SCAE is the state-of-the-art CapsNet, few research works have been done to explore the security of the SCAE. In this paper, we have confirmed the security vulnerabilities of the SCAE by an evasion attack, that is, to add perturbations to input samples so as to induce the output of the SCAE to cause misclassification [6]. Moreover, we propose a defense method called Hybrid Adversarial Training (HAT) against the above evasion attack of the SCAE.

## 3. Preliminaries

### 3.1 Stacked Capsule Autoencoder

Figure 1 shows the main structure of the SCAE. The SCAE recognizes images based on two core modules: the part capsule autoencoder (PCAE) and the object capsule autoencoder (OCAE). First, the PCAE uses a CNN to extract the pose, presence and features of each part of the objects in the input image. Then, the OCAE uses a Set Transformer [12] to encode the scattered parts obtained by the PCAE into larger objects, and output the presence probability of all objects that may appear. Finally, a classification result is given by the classifier according to the output of the OCAE.

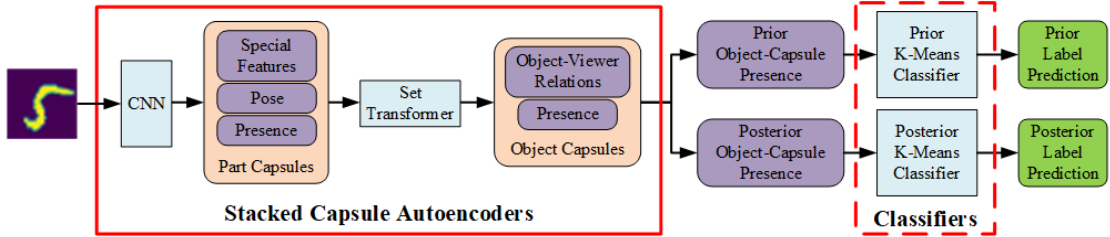

Figure 1  SCAE's architecture.

There are two types of outputs provided to the classifiers by the SCAE: the prior object-capsule presence with dimension $[B, K]$ and the posterior object-capsule presence with dimension $[B, K, M]$, where $B$ is the batch size, $K$ is the number of object capsules, and $M$ is the number of part capsules. The SCAE tells the classifiers the presence probability through these outputs whose values range from 0 to 1.

A classifier collects the outputs of the SCAE on the training dataset and chooses one type of outputs for k-means clustering. Then it uses bipartite graph matching [63] to obtain the mapping permutation between clustering labels and ground truth labels. At the testing phase, the trained k-means classifier receives the output of the SCAE and predicts the label of the input image. It is noticeable that in order to make different k-means classifiers have the same structure, the dimension $M$ of the posterior object-capsule presence is reduced and summed so as to ensure shape consistency.

### 3.2 Adversarial Training

Adversarial training is a training technique that improves the robustness of the model. In this theory, adversarial samples that can fool the model are added into the training dataset and involved in the training procedure. The model can learn the information of its vulnerabilities through these samples and enhance the ability to resist adversarial attacks. Common adversarial training methods are to add the distance between the outputs of the model on the adversarial samples and their corresponding clean ones to the loss function during the training procedure so that the trained model can obtain right outputs on adversarial samples at testing phase. Formula (1) shows the modified loss function:

$$\tilde{L}(x', \bar{y}, M) = \|\bar{y} - M(x')\|_2 \tag{1}$$

where $x$ is the clean sample, $\bar{y}$ is the one-hot encoding of the label of $x$, $x'$ is the adversarial sample that can fool the model $M$ based on $x$, and $M(x')$ means the output of $M$ on $x'$. An effective adversarial training is required to improve the robustness of the model as well as reduce the impact on the accuracy of the model as much as possible. On the other hand, the complexity of the algorithm to generate the adversarial samples for training should be limited so as to reduce the time of its computation. At present, there are studies of adversarial training such as [25,26,40–44].

### 3.3 Distillation

Proposed by Hinton et al. [64] in 2015, distillation is a technique that compresses the model while maintaining its accuracy. Usually, the more complex the model is, the stronger its fitting ability is, but the cost is that the model may become too large to be deployed on terminals. The theory of distillation is to train a complex teacher model $M_{tch}$ and then train a simplified student model $M_{stu}$ with the guidance of $M_{tch}$. The knowledge of $M_{tch}$ will be transferred to $M_{stu}$, and the two models can have similar accuracy. The key issue of distillation is the method of knowledge transfer. Hinton et al. modified the softmax function:

$$\tilde{F}(x) = \left[ \frac{e^{\frac{z_i(x)}{T}}}{\sum_{j=0}^{N-1} e^{\frac{z_j(x)}{T}}} \right]_{i \in 0..N-1} \tag{2}$$

where $x$ is the input sample, $z(x)$ is the logits output by the last hidden layer of the model, $N$ is the length of the logits vector, and $T$ is a hyper parameter called temperature. This function is called softmax with $T$. The larger the value of $T$ is, the smoother the output of the function is, and the less difficult it is to fit it as the target. During the training phase, the original softmax functions of $M_{stu}$ and $M_{tch}$ are replaced by formula (2), and the distance of the outputs of the two models on the same sample is added to the loss function of $M_{stu}$. Through this approach, the knowledge of $M_{tch}$ can be transferred to $M_{stu}$.

Adversarial distillation is a special use of distillation. During the procedure of distillation, not only is the knowledge of $M_{tch}$ learnt by $M_{stu}$, but the robustness of $M_{tch}$ is also inherited by $M_{stu}$. If $M_{stu}$ relies only on itself to improve its robustness (for example, using adversarial training), the features of adversarial samples may not be recognized by $M_{stu}$ well, or $M_{stu}$ may be overfitted on adversarial samples. Nevertheless, these problems can be mitigated by distillation. For example, combining adversarial training and distillation, or distilling $M_{stu}$ with robust $M_{tch}$, both are effective methods to train robust models [51–53].

## 4. Threat Model

In this section, we propose an evasion attack against the SCAE. The attacker constructs an adversarial perturbation and applies it to the clean sample to form an adversarial one. The adversarial sample induces the encoding process of the SCAE by decreasing the value of presence which is output by the object capsules related to the label of the clean sample so as to lower the contribution of these capsules to correct classification. The SCAE will output wrong encoding results on the adversarial sample, and the classifier will make wrong classification based on the output of the SCAE.

The SCAE uses the sparsity loss to allocate different capsules to different classes, and the object capsules related to the correct class will be activated and output high value, while the irrelevant object capsules remain inactive. The evasion attack aims at lowering the output of those activated object capsules so as to cause the SCAE to mistakenly believe that "the objects that belong to the real category do not appear in the image". This theory can be described as the following target function:

$$\text{Minimize} \quad f(x+p) = \sum_{i \in S} E(x+p)_i \tag{3}$$

where $x$ is the clean sample, $p$ is the adversarial perturbation, $E(x)_i$ means the value of presence output by the object capsule $i$ in the SCAE model $E$ on the input $x$, and $S$ is the object capsule subset related to the original category of $x$:

$$S = \left\{ i \middle| E(x)_i > \frac{1}{K} \sum_j E(x)_j \right\} \tag{4}$$

The subset $S$ includes all object capsules whose outputs are greater than the average value. According to the target function of Formula (3), we can obtain the optimization problem of the evasion attack:

$$\begin{aligned} \text{Minimize} \quad & \|p\|_2 + \alpha \cdot f(x+p) \\ s.t. \quad & x + p \in [0,1]^n \end{aligned} \tag{5}$$

where $\alpha > 0$ is a suitably chosen hyperparameter which ensures that the two parts of Formula (5) can be optimized simultaneously. As for the box constraints in Formula (5), namely, $x + p \in$

$[0,1]^n$, in order to avoid the damage of gradient propagation caused by clipping pixel values directly, we use the "change of variables" method to handle it. We map the original sample into the $arctanh$ space, compute on the mapped sample, and finally map it back to $[0,1]^n$ space. The relationship between the original and mapped samples is as follows:

$$w = \text{arctanh}(2x - 1)$$
$$p = \frac{1}{2}(\tanh(w + p') + 1) - x \quad (6)$$

where $w$ is the mapping of the original sample $x$ in the $arctanh$ space, and $p'$ is the mapping of the perturbation $p$ in the $arctanh$ space. Our algorithm computes $p'$ in the $arctanh$ space instead of $p$. The range of value does not need to be considered because the hyperbolic tangent function can map any value in $(-\infty, +\infty)$ back to $(-1,1)$. [1]

The full algorithm consists of the inner iteration and the outer iteration. In the inner iteration, we use an optimizer to solve the optimization problem of Formula (5). In the outer iteration, we initialize the optimizer, execute a complete inner iteration and update the value of $\alpha$. We perform multiple rounds of outer iterations, and select the perturbation $p$ with the smallest $\|p\|_2$ as the best result. The whole procedure is shown in Algorithm 1:

---

**Algorithm 1:** Generating the Perturbation with Optimizer

1: **Input:** Image $x$, SCAE model $E$, classifier $C$, optimizer $opt$, hyperparameter $\alpha$, the number of outer iterations $n_{o\_iter}$, the number of inner iterations $n_{i\_iter}$.
2: **Output:** Perturbation $p$.
3:
4: Initialize $p \leftarrow 0$, $\mathcal{L}_p \leftarrow +\infty$.
5: $S \leftarrow \left\{ i \middle| E(x)_i > \frac{1}{K} \sum_{i=1}^{K} E(x)_i \right\}$
6: $w \leftarrow \text{arctanh}(2x - 1)$
7: **for** $i$ in $n_{o\_iter}$ **do**
8:     Init $opt$.
9:     $p'_0 \leftarrow \text{rand}(0,1)$
10:     **for** $j$ in $n_{i\_iter}$ **do**
11:         $x_j^{adv} \leftarrow \frac{1}{2}\left(\tanh\left(w + p'_j\right) + 1\right)$
12:         $\mathcal{L} \leftarrow \left\|x_j^{adv} - x\right\|_2 + \alpha \cdot \sum_{i \in S} E\left(x_j^{adv}\right)_i$
13:         */\* The optimizer computes next $p'_j$ \*/*
14:         $p'_{j+1} \leftarrow opt\left(p'_j \middle| \mathcal{L}\right)$

---

[1] During the experiments, we use $\text{arctanh}((2x - 1) * \epsilon)$ to avoid dividing by zero.

```
15:         x_{j+1}^{adv} ← (1/2)(tanh(w + p'_{j+1}) + 1)
16:         /* Judge if the current result is the best one */
17:         if  C(E(x_{j+1}^{adv})) ≠ C(E(x))  and  ||x_{j+1}^{adv} − x||_2 < 𝓛_p  do
18:             𝓛_p ← ||x_{j+1}^{adv} − x||_2
19:             p ← x_{j+1}^{adv} − x
20:         end if
21:     end for
22:     Update α.
23: end for
24: return p
```

We set the upper bound $\alpha_{ub}$ and lower bound $\alpha_{lb}$ of $\alpha$ as $+\infty$ and 0. During the update of $\alpha$, if the algorithm obtains any adversarial sample $x_{j+1}^{adv}$ that satisfies $C(E(x_{j+1}^{adv})) \neq C(E(x))$ in the inner iterations, we let $\alpha_{ub} \leftarrow \alpha$, otherwise $\alpha_{lb} \leftarrow \alpha$, and finally take $(\alpha_{ub} + \alpha_{lb})/2$ as the new value of $\alpha$ in the next outer iteration.

We have conducted experiments on multiple datasets to verify the effectiveness of our evasion attack, and the details can be found in [6]. The experimental results show that our evasion attack can achieve high attack success rates and stealthiness, which proves the existence of security vulnerabilities in the SCAE.

## 5. The Proposed Methods to Train Robust SCAEs

In this section, we propose two basic defense models (AT-SCAE and AD-SCAE) and an advanced defense model (HAT-SCAE) against the above-mentioned threat. AT-SCAE utilizes the theory of adversarial training to promote the robustness of the SCAE. AD-SCAE further increases the accuracy of the robust SCAE model with distillation. HAT-SCAE is combination of the above two methods which not only reduces the impact on the accuracy, but also improves the defense capability of the SCAE to the evasion attacks. The goal of our defense models is to improve the resistance of the SCAE to the evasion attack while maintaining its accuracy as much as possible.

### 5.1 Training SCAE with Adversarial Training

We name the defense model with adversarial training AT-SCAE. In this method, we generate adversarial samples which can cause misclassification (see Appendix A for details) and add the generated adversarial samples to the training dataset so that the SCAE can adjust its parameters according to the features of the adversarial samples during the training phase.

The adversarial training method described in Formula (1) in section 3.2 cannot be applied to the SCAE directly because the training procedure of the SCAE is unsupervised and the labels of the samples are unavailable. On the other hand, the outputs encoded by the SCAE are the presence probabilities of different objects, and the classification results cannot be obtained until the encoding results are input into the k-means classifier.

In order to conduct adversarial training on the SCAE, we need to make use of the following mechanism in the unsupervised training.

When the SCAE encodes an image into capsules, it will also decode the capsules back into the image to make sure the capsules contain valid information about the image. This step is optimized by the reconstruction loss, which is the distance between the decoded image and the input image [1]. When conducting adversarial training, we modify the computation of the reconstruction loss. If the input image is an adversarial sample, the reconstruction loss will be the distance between the decoded image and the corresponding clean sample of the adversarial sample. During the procedure of encoding and decoding, the SCAE will discard the adversarial features and only keeps the clean ones. The loss function of AT-SCAE is as follows:

$$\mathcal{L}_{AT}(x, x', E) = \mathcal{L}_{no\_rec}(x, E) + \mathcal{L}_{rec}(x, x', E) \tag{7}$$

where $E$ is the SCAE model, $x$ is the input sample, $\mathcal{L}_{no\_rec}(x, E)$ represents all loss items in the loss function of the SCAE except the reconstruction loss, and $\mathcal{L}_{rec}(x, x', E)$ represents the reconstruction loss with the target image $x'$.

We take the input sample $x$ as the target image $x'$ of the reconstruction loss for normal training. After $k$ batches of normal training, we conduct one batch of adversarial training, that is, we take the generated adversarial samples according to Appendix A as the input samples $x$ and their corresponding clean samples as the target images $x'$ of the reconstruction loss, whereby we train the SCAE with the loss function of Formula (7) to fix potential vulnerabilities. The reason why this procedure is conducted at intervals is to mitigate the impact on the accuracy of the SCAE. The whole training procedure of AT-SCAE is shown in Algorithm 2:

| **Algorithm 2:** Training the AT-SCAE model |
|---|
| 1: **Input:** Training dataset $X$, optimizer $opt$, generator for adversarial samples $G$, the number of epochs $n_{ep}$, the number of batches before conducting one batch of adversarial training $k$. |
| 2: **Output:** AT-SCAE model $E_{AT}$. |
| 3: |
| 4: Initialize AT-SCAE model $E_{AT}$ with parameters $\theta$, $n_{bch} \leftarrow 0$. |
| 5: **for** $epoch$ in $n_{ep}$ **do** |
| 6:    **for** each batch $x$ in $X$ **do** |
| 7:       **if** $n_{bch} = k$ **do** |
| 8:          */\* Conduct one batch of adversarial training \*/* |
| 9:          $\theta \leftarrow opt(\theta \mid \mathcal{L}_{AT}(G(x), x, E_{AT}))$ |

| | | |
|---|---|---|
| 10: | | $n_{bch} \leftarrow 0$ |
| 11: | **else** | |
| 12: | | */\* Conduct  k  batches of normal training \*/* |
| 13: | | $\theta \leftarrow opt(\theta|\mathcal{L}_{AT}(x,x,E_{AT}))$ |
| 14: | | $n_{bch} \leftarrow n_{bch} + 1$ |
| 15: | **end if** | |
| 16: | **end for** | |
| 17: **end for** | | |
| 18: **return** $E_{AT}$ | | |

## 5.2 Training SCAE with Adversarial Distillation

We name the defense model with adversarial distillation AD-SCAE. This method introduces the theory of distillation to improve the accuracy of the robust SCAE model. Although AT-SCAE can enhance the robustness of the SCAE, it will decrease the accuracy of the SCAE at a certain degree. In the training procedure of AD-SCAE, we obtain a teacher model with normal training and then train a student model with the help of the teacher model based on adversarial distillation. The teacher model will guide the student model to correctly distinguish the characteristics of the adversarial samples so as to reduce their impact on the accuracy of the student model.

We conduct adversarial distillation on the output of the capsules encoded by the SCAE. Different from the traditional distillation process, AD-SCAE adds the distance between the outputs of the teacher model on the origin sample and the outputs of the student model on the adversarial sample to the loss function as the distillation loss. The knowledge of the teacher model is transferred to the student model and guides the student model to fix its vulnerabilities according to the features of the adversarial samples. As the output layer of the SCAE does not have a softmax function, there is no need to process the output with Formula (2). The modified loss function is as follows:

$$\mathcal{L}_{AD}(x, x', E_{tch}, E_{stu}, \lambda) = (1 - \lambda) * \mathcal{L}_{AT}(x, x', E_{stu}) + \lambda * \|E_{tch}(x) - E_{stu}(x')\|_2 \quad (8)$$

where $x$ is the original sample, $x'$ is the input sample for the student model (it can be the original sample or the adversarial sample), $E(x)$ represents the encoding result of the SCAE model $E$ on the input sample $x$, $E_{tch}$ and $E_{stu}$ represent the teacher model and the student model with the same structure as the SCAE, and $\lambda$ is the weight that balances the two loss items.

We conduct one batch of adversarial distillation every $k$ batches of normal distillation. Both the normal and adversarial distillation use the loss function of Formula (8) to train the SCAE. During normal distillation, $x'$ is same as the original sample $x$. When conducting adversarial distillation, we take the generated adversarial samples according to Appendix A as $x'$. The whole training procedure of AD-SCAE is shown in Algorithm 3:

**Algorithm 3:** Training the AD-SCAE model

```
 1:  Input: Training dataset X, optimizer opt, generator for adversarial samples G, pre-trained
     teacher model $E_{tch}$, hyperparameter λ, the number of epochs $n_{ep}$, the number of batches before
     conducting one batch of adversarial training k.
 2:  Output: AD-SCAE model $E_{AD}$.
 3:
 4:  Initialize AD-SCAE model $E_{AD}$ with parameters θ, $n_{bch}$ ← 0.
 5:  for epoch in $n_{ep}$ do
 6:      for each batch x in X do
 7:          if $n_{bch}$ = k do
 8:              /* Conduct one batch of adversarial distillation */
 9:              θ ← opt(θ|$\mathcal{L}_{AD}$(x, G(x), $E_{tch}$, $E_{AD}$, λ))
10:              $n_{bch}$ ← 0
11:          else
12:              /* Conduct k batches of normal distillation */
13:              θ ← opt(θ|$\mathcal{L}_{AD}$(x, x, $E_{tch}$, $E_{AD}$, λ))
14:              $n_{bch}$ ← $n_{bch}$ + 1
15:          end if
16:      end for
17:  end for
18:  return $E_{AD}$
```

## 5.3 Training SCAE with Hybrid Adversarial Training

We name the defense model with Hybrid Adversarial Training HAT-SCAE. This method is a combination of adversarial training and adversarial distillation which aims at obtaining better accuracy and robustness. AD-SCAE promotes the accuracy of the SCAE, but it has some negative effects on the robustness of the SCAE. According to the research of Zhu et al. [53], if the robustness of the teacher model for distillation is unreliable, this unreliability can be transferred to the student model and cause it have the same vulnerabilities as the teacher model has, which is also confirmed by our experimental results. In order to solve this problem, we propose the Hybrid Adversarial Training method in this section.

Adversarial distillation can promote the accuracy of the model and improve its robustness at a certain degree, while adversarial training focuses on improving the robustness of the model. Hybrid Adversarial Training means to combine adversarial distillation and adversarial training to train a more robust SCAE model. First, we use the loss function of AD-SCAE $\mathcal{L}_{AD}$ to train the SCAE model and make sure that it has enough knowledge of the clean samples and can distinguish adversarial samples at a certain degree. After the SCAE model has enough accuracy and no longer needs the guidance of a teacher model, we use the loss function of AT-SCAE $\mathcal{L}_{AT}$ to train the SCAE model in order to enhance its robustness and fix the vulnerabilities inherited from the teacher model. The whole training procedure of HAT-SCAE is shown in Algorithm 4:

| | |
|---|---|
| **Algorithm 4:** Training the HAT-SCAE model | |
| 1: | **Input:** Training dataset $X$, optimizer $opt$, generator for adversarial samples $G$, pre-trained teacher model $E_{tch}$, hyperparameter $\lambda$, the number of epochs for adversarial distillation $n_{ad}$, the number of epochs for adversarial training $n_{at}$, the number of batches before conducting one batch of adversarial distillation or adversarial training $k$. |
| 2: | **Output:** HAT-SCAE model $E_{HAT}$. |
| 3: | |
| 4: | Initialize HAT-SCAE model $E_{HAT}$ with parameters $\theta$, $n_{bch} \leftarrow 0$. |
| 5: | /* Phase 1: Adversarial Distillation */ |
| 6: | **for** $epoch$ in $n_{ad}$ **do** |
| 7: |    **for** each batch $x$ in $X$ **do** |
| 8: |       **if** $n_{bch} = k$ **do** |
| 9: |          /* Conduct one batch of adversarial distillation */ |
| 10: |          $\theta \leftarrow opt(\theta \vert \mathcal{L}_{AD}(x, G(x), E_{tch}, E_{HAT}, \lambda))$ |
| 11: |          $n_{bch} \leftarrow 0$ |
| 12: |       **else** |
| 13: |          /* Conduct $k$ batches of normal distillation */ |
| 14: |          $\theta \leftarrow opt(\theta \vert \mathcal{L}_{AD}(x, x, E_{tch}, E_{HAT}, \lambda))$ |
| 15: |          $n_{bch} \leftarrow n_{bch} + 1$ |
| 16: |       **end if** |
| 17: |    **end for** |
| 18: | **end for** |
| 19: | /* Phase 2: Adversarial Training */ |
| 20: | **for** $epoch$ in $n_{at}$ **do** |
| 21: |    **for** each batch $x$ in $X$ **do** |
| 22: |       **if** $n_{bch} = k$ **do** |
| 23: |          /* Conduct one batch of adversarial training */ |
| 24: |          $\theta \leftarrow opt(\theta \vert \mathcal{L}_{AT}(x, G(x), E_{HAT}))$ |
| 25: |          $n_{bch} \leftarrow 0$ |
| 26: |       **else** |
| 27: |          /* Conduct $k$ batches of normal training */ |
| 28: |          $\theta \leftarrow opt(\theta \vert \mathcal{L}_{AT}(x, x, E_{HAT}))$ |
| 29: |          $n_{bch} \leftarrow n_{bch} + 1$ |
| 30: |       **end if** |
| 31: |    **end for** |
| 32: | **end for** |
| 33: | **return** $E_{HAT}$ |

## 6. Experimental Evaluation

In the experiments, we train the SCAE models with the defense methods proposed in section 5 separately, test their robustness against the evasion attack method proposed in section 4, and compare their classification accuracy on clean and adversarial samples.

## 6.1 Experimental Setup

**Baseline:**

As there exists few researches dedicated on improving the security robustness of the SCAE, and as both adversarial training and adversarial distillation are widely used to defend against adversarial attacks on neural network models, we use AT-SCAE and AD-SCAE as the baseline to show the advantage of our defense model, namely, HAT-SCAE. We train these models separately and make a comparison of their classification accuracy (the ratio of the number of samples that are classified correctly by the classifier to the total number of all test samples) on clean and adversarial samples respectively.

**Datasets:**

Because distillation has certain requirements for the accuracy of the teacher model, when applying the defense methods to datasets that are difficult to be identified by the SCAE, the improvement of robustness will be limited. The experimental results given by Kosiorek et al. in the original paper of the SCAE [1] show that although the SCAE can achieve 98.7% accuracy on the MNIST dataset, it can only achieve 55.33% and 25.01% accuracy on the SVHN and CIFAR10 datasets, which does not satisfy the need of our experiments. ImageNet, which is similar to CIFAR10, is also not suitable for our experiments. After comparing the performance of the SCAE on various datasets, we select the MNIST and Fashion-MNIST datasets for experiments in order to show the influence of different defense methods to the robustness of the SCAE.

**Parameters:**

We train the SCAE models on the two datasets with the parameters shown in Table 1 [1,65]:

Table 1  SCAEs' main parameters.

| Dataset | MNIST | Fashion MNIST |
|---|---|---|
| Canvas size | 40 | 40 |
| Num of part capsules | 24 | 24 |
| Num of object capsules | 24 | 24 |
| Num of channels | 1 | 1 |
| Template size | 11 | 11 |
| Part capsule noise scale | 4.0 | 4.0 |
| Object capsule noise scale | 4.0 | 4.0 |
| Part CNN | 2×(128:2)-2×(128:1) | 2×(128:2)-2×(128:1) |
| Set Transformer | 3×(1-16)-256 | 3×(1-16)-256 |

The main parameters are the same as Kosiorek et al. used in the original paper of the SCAE [1]. For part CNN, 2×(128:2) means two convolutional layers with 128 channels and a stride of two. For the set transformer, 3×(1-16)-256 means three layers, one attention head, 16 hidden units and 256 output units. All models use the same optimizer for training. The main parameters of the

optimizer are shown in Table 2:

Table 2　Settings of the optimizer to train the SCAEs.

| Optimizer parameter | Value |
|---|---|
| Algorithm | RMSProp |
| Learning rate | $3\times10^{-5}$ |
| Momentum | 0.9 |
| $\epsilon$ | $1\times10^{-6}$ |
| Learning rate decay steps | 10000 |
| Learning rate decay rate | 0.96 |
| Batch size | 100 |

According to the information given by Kosiorek et al. [1], the values of $k$ of all k-means classifiers are set as 10, which is the same as the number of categories of the datasets.

Before the defense experiments, the teacher models for distillation need to be pre-trained. The accuracy of all teacher models is shown in Table 3:

Table 3　Accuracy of the pre-trained teacher model in %.

| Dataset | MNIST | Fashion MNIST |
|---|---|---|
| Prior k-means classifier | 97.40 | 67.71 |
| Posterior k-means classifier | 97.63 | 66.83 |

## 6.2 Experimental Method

Our experiments consist of two steps:

1. We train the AT-SCAE, AD-SCAE and HAT-SCAE models separately, and evaluate the classification accuracy of the models on clean samples;
2. We attack the prior k-means classifiers and the posterior k-means classifiers with the evasion attack method proposed in section 4. In each attack experiment, we randomly choose 5000 samples and generate adversarial samples based on them. Then we input the adversarial samples to the SCAE model to obtain the encoding results. The encoding results are then input into the classifier to get the classification results. We evaluate the classification accuracy of the models on adversarial samples.

The source code of our defense experiments is available at https://github.com/FrostbiteXSW/SCAE_Defense.

In the experiments, the number of epochs of training $n_{ep}$ in AT-SCAE and AD-SCAE is set as 100. The number of epochs of adversarial distillation $n_{ad}$ and the number of epochs of adversarial training $n_{at}$ in HAT-SCAE are set as 50. The number of interval batches between two batches of adversarial training and adversarial distillation $k$ in all defense methods is set as 1. The weight value $\lambda$ in Formula (8) is set as 0.5.

For the generation algorithm of adversarial samples for training, we set the numbers of inner and outer iterations as 5 and 30 respectively. The hyperparameter $\beta$ in Formula (9) is set as 1. The initial value of hyperparameter $\alpha$ is set as 100, and the upper and lower bounds of $\alpha$ are set as $+\infty$ and 0. If the upper bound is $+\infty$ when updating $\alpha$, let $\alpha \leftarrow \alpha \times 10$. The target classifier of the generation algorithm of adversarial samples for training is the built-in posterior linear classifier of the SCAE [1] instead of the k-means classifiers, and the reason for this is to remove the time cost of updating the k-means classifiers repeatedly so as to accelerate the training procedure.

For the evasion attack algorithm for robustness evaluation, we set the numbers of inner and outer iterations as 9 and 300 respectively. The initial value of hyperparameter $\alpha$ is set as 100, and the upper and lower bounds of $\alpha$ are set as $+\infty$ and 0. If the upper bound is $+\infty$ when updating $\alpha$, let $\alpha \leftarrow \alpha \times 10$. The main parameters of the optimizer used by the evasion attack algorithm are shown in Table 4:

Table 4  Settings of the optimizer used to attack the SCAEs.

| Optimizer parameter | Value |
| --- | --- |
| Algorithm | Adam |
| Learning rate | 1.0 |
| $\beta_1$ | 0.9 |
| $\beta_2$ | 0.999 |
| $\epsilon$ | 1x10$^{-8}$ |

The attack success rate increases with the amount of perturbation, while the stealthiness becomes worse. The amount of perturbation needs to be limited in order to objectively evaluate the effect of different defense methods. In our experiments, the thresholds of the $L_2$ norms of the perturbations on the MNIST and Fashion-MNIST datasets are set as 4 and 5 respectively, which ensure that the adversarial perturbations are imperceptible.

## 6.3 Results and Discussion

First, we train the AT-SCAE, AD-SCAE and HAT-SCAE models on the MNIST and Fashion-MNIST datasets, then construct the prior k-means classifiers and the posterior k-means classifiers for each model separately, and finally evaluate the classification accuracy of the models on clean samples, which is shown in Table 5:

Table 5  The comparison of classification accuracy of the original SCAE and the three robust models on clean samples.

| Classifier | Defense Model | MNIST (%) | Fashion-MNIST (%) |
|---|---|---|---|
| Prior k-means classifier | SCAE | 97.40 | **67.71** |
| | AT-SCAE | 85.18 | 58.87 |
| | AD-SCAE | **97.68** | 61.15 |
| | HAT-SCAE | 95.22 | 59.17 |
| Posterior k-means classifier | SCAE | 97.63 | **66.83** |
| | AT-SCAE | 80.97 | 59.64 |
| | AD-SCAE | **97.92** | 61.03 |
| | HAT-SCAE | 95.85 | 60.98 |

Among the three defense models, AD-SCAE utilizes the knowledge of the teacher model and obtains the best accuracy. As there is no teacher model in the training procedure of AT-SCAE, the accuracy of the model is the worst. HAT-SCAE is a combination of AT-SCAE and AD-SCAE, and the accuracy of its model is between them as well.

After getting the three robust models, we use the evasion attack algorithm proposed in section 4 to attack the prior k-means classifiers and the posterior k-means classifiers on the MNIST and Fashion-MNIST datasets respectively, and evaluate the classification accuracy of the models on adversarial samples, which is shown in Table 6:

Table 6    The comparison of classification accuracy of the original SCAE and the three robust models on adversarial samples.

| Classifier | Defense Model | MNIST (%) | Fashion-MNIST (%) |
|---|---|---|---|
| Prior k-means classifier | SCAE | 0.38 | 3.82 |
| | AT-SCAE | 58.56 | 20.70 |
| | AD-SCAE | 66.80 | 36.20 |
| | HAT-SCAE | **81.36** | **37.38** |
| Posterior k-means classifier | SCAE | 0.76 | 1.86 |
| | AT-SCAE | 64.50 | **37.94** |
| | AD-SCAE | 58.26 | 25.62 |
| | HAT-SCAE | **82.14** | 37.20 |

The above results show that all three defense models can achieve better robustness. HAT-SCAE performs the best, as the promotion of classification accuracy on adversarial samples of the MNIST dataset is much greater than the other two defense models. The experimental results on the Fashion-MNIST dataset also prove the effectiveness of the three defense models. As the SCAE is not capable enough to fit the Fashion-MNIST dataset well, the effect of the defense models is also limited.

In order to clearly describe the effect of the defense models, taking the MNIST dataset as an example, we draw the variation diagram of the classification accuracy with the perturbation amount threshold and analyze the variation of robustness with the increase of the amount of perturbation:

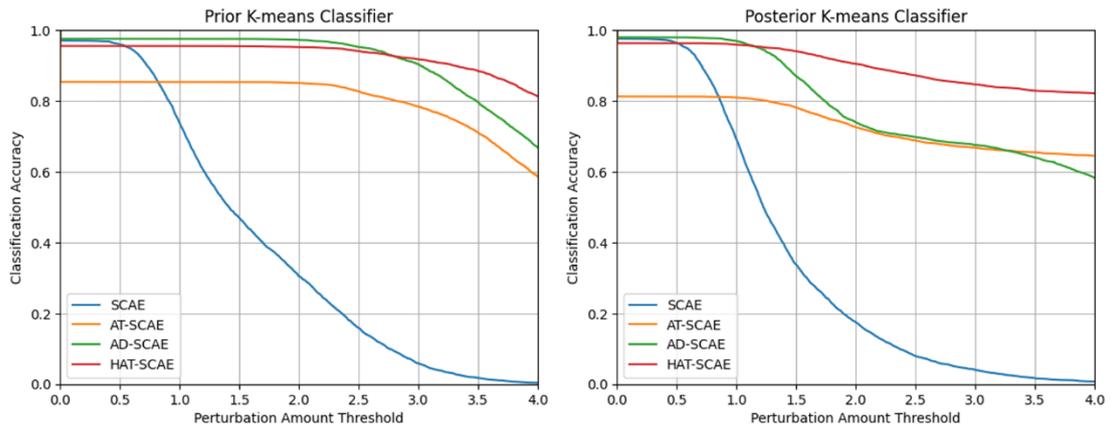

Figure 2  The variation of classification accuracy with the increase of the amount of perturbation on the MNIST dataset. Each coordinate point $(x, y)$ represents that when the perturbation amount of each adversarial sample is not greater than $x$, the classification accuracy of the model on these adversarial samples equals $y$. The cross point of the curve and the y-axis represents the classification accuracy of the model on clean samples.

It is shown in Figure 2 that the three defense models can improve the defense capability of the SCAE to evasion attacks on the MNIST dataset. AT-SCAE trains model on adversarial samples and has a negative effect on the model accuracy on clean samples. Due to the advantage of distillation, AD-SCAE significantly improves the model accuracy on clean samples, but the cost is that the vulnerabilities in the teacher model are transferred to the student model and weaken its robustness at a certain degree. HAT-SCAE has better classification accuracy on both clean and adversarial samples, which proves HAT to be the best one among the three defense methods.

The variation diagram of the classification accuracy with the perturbation amount threshold on the Fashion-MNIST dataset is shown in Figure 3:

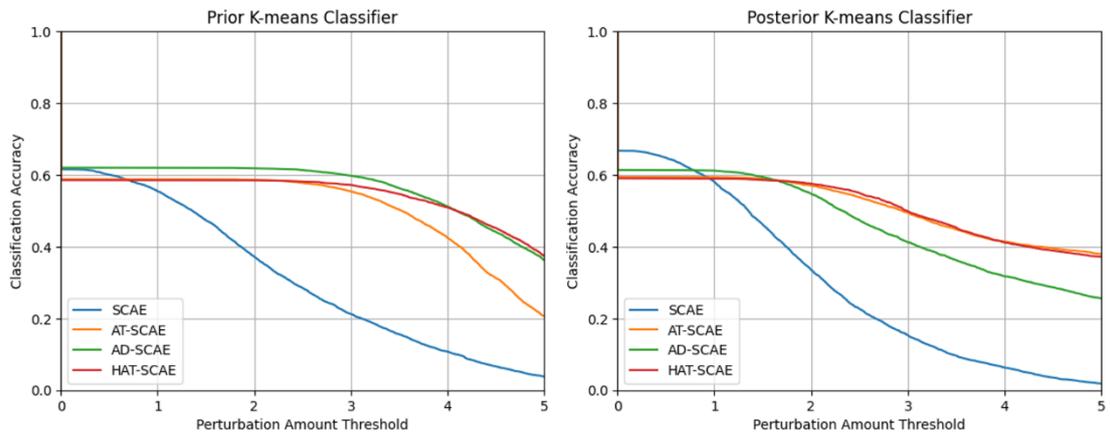

Figure 3  The variation of classification accuracy with the increase of the amount of perturbation on the Fashion-MNIST dataset.

The difference of classification accuracy of the three defense models on the Fashion-MNIST dataset is not as obvious as that on the MNIST dataset. This is caused by the fitting ability of the SCAE. As adversarial training will decrease the accuracy of the model, we introduce the theory of distillation to optimize the training procedure, and the effect of distillation increases with the performance of the teacher model. It can be seen from Table 3 that the classification accuracy of the original SCAE model on the Fashion-MNIST dataset is worse than that on the MNIST dataset, so is the effect of the teacher model. Under this circumstance, the three defense models have similar classification accuracy on clean samples, but with the increase of the amount of perturbation, the robustness of AT-SCAE and AD-SCAE decreases while HAT-SCAE's performance is relatively stable, which is also the best one among the three defense models.

### 6.4 Ablation Study

In the training procedure of HAT-SCAE, the first step is adversarial distillation, and the second step is adversarial training. We provide an ablation study in this section, which replace the first step with normal training so as to study the influence of adversarial distillation to the robustness of HAT-SCAE. We name the SCAE model obtained by the above method as NTAT-SCAE (Stacked Capsule Autoencoder with Normal Training and Adversarial Training).

We train the NTAT-SCAE model in the same experimental method and construct the prior k-means classifier and the posterior k-means classifier for it. The classification accuracy on clean samples is shown in Table 7:

Table 7  The comparison of classification accuracy on clean samples between NTAT-SCAE and HAT-SCAE.

| Classifier | Defense Model | MNIST (%) | Fashion-MNIST (%) |
|---|---|---|---|
| Prior k-means classifier | SCAE | **97.40** | **67.71** |
|  | NTAT-SCAE | 96.74 | 60.11 |
|  | HAT-SCAE | 95.22 | 59.17 |
| Posterior k-means classifier | SCAE | **97.63** | **66.83** |
|  | NTAT-SCAE | 96.94 | 61.74 |
|  | HAT-SCAE | 95.85 | 60.98 |

As normal training does not add adversarial samples to the training dataset, the classification accuracy of the model trained by NTAT-SCAE is somewhat higher than that of the model trained by HAT-SCAE. We attack the model with the above evasion attack method on the MNIST and Fashion-MNIST datasets to test its robustness, and the results are shown in Table 8:

Table 8  The comparison of classification accuracy on adversarial samples between NTAT-SCAE and HAT-SCAE.

| Classifier | Defense Model | MNIST (%) | Fashion-MNIST (%) |
|---|---|---|---|
| Prior k-means classifier | SCAE | 0.38 | 3.82 |
| | NTAT-SCAE | 42.20 | 33.26 |
| | HAT-SCAE | **81.36** | **37.38** |
| Posterior k-means classifier | SCAE | 0.76 | 1.86 |
| | NTAT-SCAE | 51.90 | 33.92 |
| | HAT-SCAE | **82.14** | **37.20** |

From the above results it can be observed that the robustness of NTAT-SCAE is worse than HAT-SCAE. In order to clearly show the gap between the two defense models, we draw the variation diagram of the classification accuracy with the perturbation amount threshold for SCAE, HAT-SCAE and NTAT-SCAE:

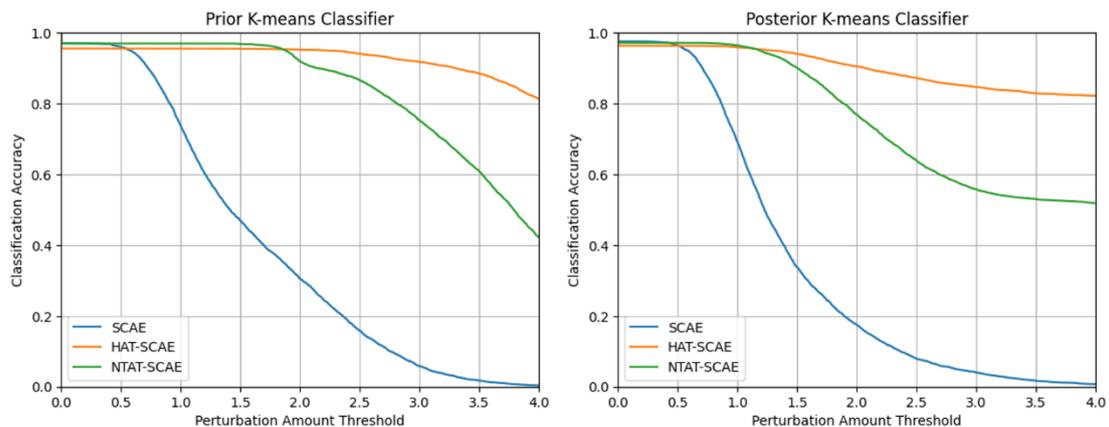

Figure 4    The variation of classification accuracy with the increase of the amount of perturbation of SCAE, HAT-SCAE and NTAT-SCAE on the MNIST dataset.

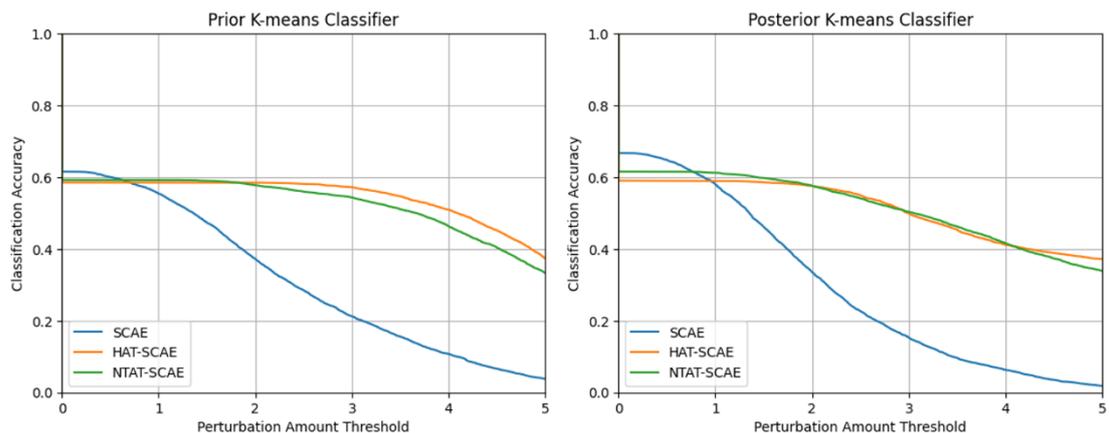

Figure 5    The variation of classification accuracy with the increase of the amount of perturbation of SCAE, HAT-SCAE and NTAT-SCAE on the Fashion-MNIST dataset.

Figure 4 and Figure 5 show the advantage of adversarial distillation on the robustness of the SCAE model. The experimental results of HAT-SCAE on the MNIST dataset is obviously better than that of NTAT-SCAE. On the Fashion-MNIST dataset, as the classification accuracy of the teacher model is imperfect, the performance of HAT-SCAE is slightly better than that of NTAT-SCAE. From the above results, we can draw a conclusion that although the classification accuracy on clean samples is similar between the two defense models, the robustness of HAT-SCAE is better than that of NTAT-SCAE, which proves that adversarial distillation is necessary for HAT-SCAE to improve the robustness of the SCAE model.

## 7. Conclusion

In this paper, we propose three models to defense evasion attacks on the SCAE: AT-SCAE, AD-SCAE and HAT-SCAE. AT-SCAE improves the robustness of the SCAE based on adversarial training. AD-SCAE further introduces the theory of distillation to reduce the negative impact of adversarial training on the accuracy of the SCAE. HAT-SCAE combines the advantage of AT-SCAE and AD-SCAE, which not only ensures that the classification accuracy of the robust model is similar to that of the plain model, but also enhances the robustness of the SCAE to evasion attacks. We evaluate the performance of the defense models with the experiments, and the experimental results show that all three defense models can improve the robustness of the SCAE. HAT-SCAE, which has less impact on the accuracy of the model and achieve the best robustness, is the best one among the three defense models to enhance the robustness of the SCAE. In our future work, we will further improve the robustness of the SCAE and explore the application of the SCAE to security domain.

# References

[1] A.R. Kosiorek, S. Sabour, Y.W. Teh, G.E. Hinton, Stacked Capsule Autoencoders, in: H.M. Wallach, H. Larochelle, A. Beygelzimer, F. d'Alché-Buc, E.B. Fox, R. Garnett (Eds.), Adv. Neural Inf. Process. Syst. 32 Annu. Conf. Neural Inf. Process. Syst. 2019 NeurIPS 2019 Dec. 8-14 2019 Vanc. BC Can., 2019: pp. 15486–15496. https://proceedings.neurips.cc/paper/2019/hash/2e0d41e02c5be4668ec1b0730b3346a8-Abstract.html.

[2] J. Yoon, Adversarial Attack to Capsule Networks, (2017). https://github.com/jaesik817/adv_attack_capsnet (accessed March 11, 2021).

[3] F. Michels, T. Uelwer, E. Upschulte, S. Harmeling, On the Vulnerability of Capsule Networks to Adversarial Attacks, CoRR. abs/1906.03612 (2019). http://arxiv.org/abs/1906.03612.

[4] A. Marchisio, G. Nanfa, F. Khalid, M.A. Hanif, M. Martina, M. Shafique, CapsAttacks: Robust and Imperceptible Adversarial Attacks on Capsule Networks, CoRR. abs/1901.09878 (2019). http://arxiv.org/abs/1901.09878.

[5] A. De Marco, Capsule Networks Robustness against Adversarial Attacks and Affine Transformations, PhD Thesis, Politecnico di Torino, 2020. https://webthesis.biblio.polito.it/14429/1/tesi.pdf.

[6] J. Dai, S. Xiong, An Evasion Attack against Stacked Capsule Autoencoder, ArXiv201007230 Cs. (2021). http://arxiv.org/abs/2010.07230 (accessed June 24, 2021).

[7] Y. Li, H. Su, J. Zhu, AdvCapsNet: To defense adversarial attacks based on Capsule networks, J Vis Commun Image Represent. 75 (2021) 103037. https://doi.org/10.1016/j.jvcir.2021.103037.

[8] D. Peer, S. Stabinger, A.J. Rodríguez-Sánchez, Increasing the adversarial robustness and explainability of capsule networks with $\backslash gamma$-capsules, CoRR. abs/1812.09707 (2018). http://arxiv.org/abs/1812.09707.

[9] S. Garg, J. Alexander, T. Kothari, Using Capsule Networks with Thermometer Encoding to Defend Against Adversarial Attacks, Proc. CS229 Final Proj. Sess. Stanf. CA. (2017).

[10] S. Sabour, N. Frosst, G.E. Hinton, Dynamic Routing Between Capsules, in: I. Guyon, U. von Luxburg, S. Bengio, H.M. Wallach, R. Fergus, S.V.N. Vishwanathan, R. Garnett (Eds.), Adv. Neural Inf. Process. Syst. 30 Annu. Conf. Neural Inf. Process. Syst. 2017 Dec. 4-9 2017 Long Beach CA USA, 2017: pp. 3856–3866. https://proceedings.neurips.cc/paper/2017/hash/2cad8fa47bbef282badbb8de5374b894-Abstract.html.

[11] G.E. Hinton, S. Sabour, N. Frosst, Matrix capsules with EM routing, in: 6th Int. Conf. Learn. Represent. ICLR 2018 Vanc. BC Can. April 30 - May 3 2018 Conf. Track Proc., OpenReview.net, 2018. https://openreview.net/forum?id=HJWLfGWRb.

[12] J. Lee, Y. Lee, J. Kim, A.R. Kosiorek, S. Choi, Y.W. Teh, Set Transformer: A Framework for Attention-based Permutation-Invariant Neural Networks, in: K. Chaudhuri, R. Salakhutdinov (Eds.), Proc. 36th Int. Conf. Mach. Learn. ICML 2019 9-15 June 2019 Long Beach Calif. USA, PMLR, 2019: pp. 3744–3753. http://proceedings.mlr.press/v97/lee19d.html.

[13] X. Chen, C. Liu, B. Li, K. Lu, D. Song, Targeted Backdoor Attacks on Deep Learning Systems Using Data Poisoning, CoRR. abs/1712.05526 (2017). http://arxiv.org/abs/1712.05526.

[14] H. Zhong, C. Liao, A.C. Squicciarini, S. Zhu, D.J. Miller, Backdoor Embedding in

Convolutional Neural Network Models via Invisible Perturbation, in: V. Roussev, B.M. Thuraisingham, B. Carminati, M. Kantarcioglu (Eds.), CODASPY 20 Tenth ACM Conf. Data Appl. Secur. Priv. New Orleans USA March 16-18 2020, ACM, 2020: pp. 97–108. https://doi.org/10.1145/3374664.3375751.

[15] A. Shafahi, W.R. Huang, M. Najibi, O. Suciu, C. Studer, T. Dumitras, T. Goldstein, Poison Frogs! Targeted Clean-Label Poisoning Attacks on Neural Networks, in: S. Bengio, H.M. Wallach, H. Larochelle, K. Grauman, N. Cesa-Bianchi, R. Garnett (Eds.), Adv. Neural Inf. Process. Syst. 31 Annu. Conf. Neural Inf. Process. Syst. 2018 NeurIPS 2018 Dec. 3-8 2018 Montr. Can., 2018: pp. 6106–6116. https://proceedings.neurips.cc/paper/2018/hash/22722a343513ed45f14905eb07621686-Abstract.html.

[16] A. Saha, A. Subramanya, H. Pirsiavash, Hidden Trigger Backdoor Attacks, in: Thirty-Fourth AAAI Conf. Artif. Intell. AAAI 2020 Thirty-Second Innov. Appl. Artif. Intell. Conf. IAAI 2020 Tenth AAAI Symp. Educ. Adv. Artif. Intell. EAAI 2020 N. Y. NY USA Febr. 7-12 2020, AAAI Press, 2020: pp. 11957–11965. https://aaai.org/ojs/index.php/AAAI/article/view/6871.

[17] J. Dai, C. Chen, Y. Li, A backdoor attack against LSTM-based text classification systems, IEEE Access. 7 (2019) 138872–138878.

[18] Y. Yao, H. Li, H. Zheng, B.Y. Zhao, Latent Backdoor Attacks on Deep Neural Networks, in: L. Cavallaro, J. Kinder, X. Wang, J. Katz (Eds.), Proc. 2019 ACM SIGSAC Conf. Comput. Commun. Secur. CCS 2019 Lond. UK Novemb. 11-15 2019, ACM, 2019: pp. 2041–2055. https://doi.org/10.1145/3319535.3354209.

[19] J. Shen, X. Zhu, D. Ma, TensorClog: An Imperceptible Poisoning Attack on Deep Neural Network Applications, IEEE Access. 7 (2019) 41498–41506. https://doi.org/10.1109/ACCESS.2019.2905915.

[20] C. Zhu, W.R. Huang, H. Li, G. Taylor, C. Studer, T. Goldstein, Transferable Clean-Label Poisoning Attacks on Deep Neural Nets, in: K. Chaudhuri, R. Salakhutdinov (Eds.), Proc. 36th Int. Conf. Mach. Learn. ICML 2019 9-15 June 2019 Long Beach Calif. USA, PMLR, 2019: pp. 7614–7623. http://proceedings.mlr.press/v97/zhu19a.html.

[21] Y. Liu, S. Ma, Y. Aafer, W.-C. Lee, J. Zhai, W. Wang, X. Zhang, Trojaning Attack on Neural Networks, in: 25th Annu. Netw. Distrib. Syst. Secur. Symp. NDSS 2018 San Diego Calif. USA Febr. 18-21 2018, The Internet Society, 2018. http://wp.internetsociety.org/ndss/wp-content/uploads/sites/25/2018/02/ndss2018_03A-5_Liu_paper.pdf.

[22] H. Kwon, H. Yoon, K.-W. Park, Selective Poisoning Attack on Deep Neural Networks †, Symmetry. 11 (2019) 892. https://doi.org/10.3390/sym11070892.

[23] N. Akhtar, A.S. Mian, Threat of Adversarial Attacks on Deep Learning in Computer Vision: A Survey, IEEE Access. 6 (2018) 14410–14430. https://doi.org/10.1109/ACCESS.2018.2807385.

[24] C. Szegedy, W. Zaremba, I. Sutskever, J. Bruna, D. Erhan, I.J. Goodfellow, R. Fergus, Intriguing properties of neural networks, in: Y. Bengio, Y. LeCun (Eds.), 2nd Int. Conf. Learn. Represent. ICLR 2014 Banff AB Can. April 14-16 2014 Conf. Track Proc., 2014. http://arxiv.org/abs/1312.6199.

[25] I.J. Goodfellow, J. Shlens, C. Szegedy, Explaining and Harnessing Adversarial Examples, in: Y. Bengio, Y. LeCun (Eds.), 3rd Int. Conf. Learn. Represent. ICLR 2015 San Diego CA USA May 7-9 2015 Conf. Track Proc., 2015. http://arxiv.org/abs/1412.6572.

[26] S.-M. Moosavi-Dezfooli, A. Fawzi, P. Frossard, DeepFool: A Simple and Accurate Method to Fool Deep Neural Networks, in: 2016 IEEE Conf. Comput. Vis. Pattern Recognit. CVPR 2016 Las Vegas NV USA June 27-30 2016, IEEE Computer Society, 2016: pp. 2574–2582. https://doi.org/10.1109/CVPR.2016.282.

[27] S.-M. Moosavi-Dezfooli, A. Fawzi, O. Fawzi, P. Frossard, Universal Adversarial Perturbations, in: 2017 IEEE Conf. Comput. Vis. Pattern Recognit. CVPR 2017 Honol. HI USA July 21-26 2017, IEEE Computer Society, 2017: pp. 86–94. https://doi.org/10.1109/CVPR.2017.17.

[28] N. Carlini, D.A. Wagner, Towards Evaluating the Robustness of Neural Networks, in: 2017 IEEE Symp. Secur. Priv. SP 2017 San Jose CA USA May 22-26 2017, IEEE Computer Society, 2017: pp. 39–57. https://doi.org/10.1109/SP.2017.49.

[29] A. Kurakin, I.J. Goodfellow, S. Bengio, Adversarial examples in the physical world, in: 5th Int. Conf. Learn. Represent. ICLR 2017 Toulon Fr. April 24-26 2017 Workshop Track Proc., OpenReview.net, 2017. https://openreview.net/forum?id=HJGU3Rodl.

[30] J. Su, D.V. Vargas, K. Sakurai, One Pixel Attack for Fooling Deep Neural Networks, IEEE Trans Evol Comput. 23 (2019) 828–841. https://doi.org/10.1109/TEVC.2019.2890858.

[31] S. Sarkar, A. Bansal, U. Mahbub, R. Chellappa, UPSET and ANGRI : Breaking High Performance Image Classifiers, CoRR. abs/1707.01159 (2017). http://arxiv.org/abs/1707.01159.

[32] S. Baluja, I. Fischer, Adversarial Transformation Networks: Learning to Generate Adversarial Examples, CoRR. abs/1703.09387 (2017). http://arxiv.org/abs/1703.09387.

[33] M. Cissé, Y. Adi, N. Neverova, J. Keshet, Houdini: Fooling Deep Structured Prediction Models, CoRR. abs/1707.05373 (2017). http://arxiv.org/abs/1707.05373.

[34] S.U. Din, N. Akhtar, S. Younis, F. Shafait, A. Mansoor, M. Shafique, Steganographic universal adversarial perturbations, Pattern Recognit Lett. 135 (2020) 146–152. https://doi.org/10.1016/j.patrec.2020.04.025.

[35] W. Brendel, J. Rauber, M. Bethge, Decision-Based Adversarial Attacks: Reliable Attacks Against Black-Box Machine Learning Models, in: 6th Int. Conf. Learn. Represent. ICLR 2018 Vanc. BC Can. April 30 - May 3 2018 Conf. Track Proc., OpenReview.net, 2018. https://openreview.net/forum?id=SyZI0GWCZ.

[36] N. Papernot, P.D. McDaniel, S. Jha, M. Fredrikson, Z.B. Celik, A. Swami, The Limitations of Deep Learning in Adversarial Settings, in: IEEE Eur. Symp. Secur. Priv. EuroSP 2016 Saarbr. Ger. March 21-24 2016, IEEE, 2016: pp. 372–387. https://doi.org/10.1109/EuroSP.2016.36.

[37] H. Xu, Y. Ma, H. Liu, D. Deb, H. Liu, J. Tang, A.K. Jain, Adversarial Attacks and Defenses in Images, Graphs and Text: A Review, Int J Autom Comput. 17 (2020) 151–178. https://doi.org/10.1007/s11633-019-1211-x.

[38] S. Rifai, P. Vincent, X. Muller, X. Glorot, Y. Bengio, Contractive Auto-Encoders: Explicit Invariance During Feature Extraction, in: L. Getoor, T. Scheffer (Eds.), Proc. 28th Int. Conf. Mach. Learn. ICML 2011 Bellevue Wash. USA June 28 - July 2 2011, Omnipress, 2011: pp. 833–840. https://icml.cc/2011/papers/455\_icmlpaper.pdf.

[39] N. Papernot, P.D. McDaniel, X. Wu, S. Jha, A. Swami, Distillation as a Defense to Adversarial Perturbations Against Deep Neural Networks, in: IEEE Symp. Secur. Priv. SP 2016 San Jose CA USA May 22-26 2016, IEEE Computer Society, 2016: pp. 582–597. https://doi.org/10.1109/SP.2016.41.

[40] A. Madry, A. Makelov, L. Schmidt, D. Tsipras, A. Vladu, Towards Deep Learning Models Resistant to Adversarial Attacks, in: 6th Int. Conf. Learn. Represent. ICLR 2018 Vanc. BC Can. April 30 - May 3 2018 Conf. Track Proc., OpenReview.net, 2018. https://openreview.net/forum?id=rJzIBfZAb.

[41] F. Tramèr, A. Kurakin, N. Papernot, I.J. Goodfellow, D. Boneh, P.D. McDaniel, Ensemble Adversarial Training: Attacks and Defenses, in: 6th Int. Conf. Learn. Represent. ICLR 2018 Vanc. BC Can. April 30 - May 3 2018 Conf. Track Proc., OpenReview.net, 2018. https://openreview.net/forum?id=rkZvSe-RZ.

[42] A. Sinha, H. Namkoong, J.C. Duchi, Certifying Some Distributional Robustness with Principled Adversarial Training, in: 6th Int. Conf. Learn. Represent. ICLR 2018 Vanc. BC Can. April 30 - May 3 2018 Conf. Track Proc., OpenReview.net, 2018. https://openreview.net/forum?id=Hk6kPgZA-.

[43] A. Shafahi, M. Najibi, A. Ghiasi, Z. Xu, J.P. Dickerson, C. Studer, L.S. Davis, G. Taylor, T. Goldstein, Adversarial training for free!, in: H.M. Wallach, H. Larochelle, A. Beygelzimer, F. d'Alché-Buc, E.B. Fox, R. Garnett (Eds.), Adv. Neural Inf. Process. Syst. 32 Annu. Conf. Neural Inf. Process. Syst. 2019 NeurIPS 2019 Dec. 8-14 2019 Vanc. BC Can., 2019: pp. 3353–3364. https://proceedings.neurips.cc/paper/2019/hash/7503cfacd12053d309b6bed5c89de212-Abstract.html.

[44] T. Miyato, A.M. Dai, I.J. Goodfellow, Adversarial Training Methods for Semi-Supervised Text Classification, in: 5th Int. Conf. Learn. Represent. ICLR 2017 Toulon Fr. April 24-26 2017 Conf. Track Proc., OpenReview.net, 2017. https://openreview.net/forum?id=r1X3g2\_xl.

[45] J. Buckman, A. Roy, C. Raffel, I.J. Goodfellow, Thermometer Encoding: One Hot Way To Resist Adversarial Examples, in: 6th Int. Conf. Learn. Represent. ICLR 2018 Vanc. BC Can. April 30 - May 3 2018 Conf. Track Proc., OpenReview.net, 2018. https://openreview.net/forum?id=S18Su--CW.

[46] C. Guo, M. Rana, M. Cissé, L. van der Maaten, Countering Adversarial Images using Input Transformations, in: 6th Int. Conf. Learn. Represent. ICLR 2018 Vanc. BC Can. April 30 - May 3 2018 Conf. Track Proc., OpenReview.net, 2018. https://openreview.net/forum?id=SyJ7ClWCb.

[47] G.S. Dhillon, K. Azizzadenesheli, Z.C. Lipton, J. Bernstein, J. Kossaifi, A. Khanna, A. Anandkumar, Stochastic Activation Pruning for Robust Adversarial Defense, in: 6th Int. Conf. Learn. Represent. ICLR 2018 Vanc. BC Can. April 30 - May 3 2018 Conf. Track Proc., OpenReview.net, 2018. https://openreview.net/forum?id=H1uR4GZRZ.

[48] C. Xie, J. Wang, Z. Zhang, Z. Ren, A.L. Yuille, Mitigating Adversarial Effects Through Randomization, in: 6th Int. Conf. Learn. Represent. ICLR 2018 Vanc. BC Can. April 30 - May 3 2018 Conf. Track Proc., OpenReview.net, 2018. https://openreview.net/forum?id=Sk9yuql0Z.

[49] T.L. Le, N.P. Park, D. Lee, A Sweet Rabbit Hole by DARCY: Using Honeypots to Detect Universal Trigger's Adversarial Attacks, in: 59th Annu. Meet. Assoc. Comp Linguist. ACL, 2021.

[50] S. Shan, E. Wenger, B. Wang, B. Li, H. Zheng, B.Y. Zhao, Gotta Catch'Em All: Using Honeypots to Catch Adversarial Attacks on Neural Networks, in: J. Ligatti, X. Ou, J. Katz, G. Vigna (Eds.), CCS 20 2020 ACM SIGSAC Conf. Comput. Commun. Secur. Virtual Event

USA Novemb. 9-13 2020, ACM, 2020: pp. 67–83. https://doi.org/10.1145/3372297.3417231.

[51] T. Chen, Z. Zhang, S. Liu, S. Chang, Z. Wang, Robust Overfitting may be mitigated by properly learned smoothening, in: 9th Int. Conf. Learn. Represent. ICLR 2021 Virtual Event Austria May 3-7 2021, OpenReview.net, 2021. https://openreview.net/forum?id=qZzy5urZw9.

[52] M. Goldblum, L. Fowl, S. Feizi, T. Goldstein, Adversarially Robust Distillation, in: Thirty-Fourth AAAI Conf. Artif. Intell. AAAI 2020 Thirty-Second Innov. Appl. Artif. Intell. Conf. IAAI 2020 Tenth AAAI Symp. Educ. Adv. Artif. Intell. EAAI 2020 N. Y. NY USA Febr. 7-12 2020, AAAI Press, 2020: pp. 3996–4003. https://aaai.org/ojs/index.php/AAAI/article/view/5816.

[53] J. Zhu, J. Yao, B. Han, J. Zhang, T. Liu, G. Niu, J. Zhou, J. Xu, H. Yang, Reliable Adversarial Distillation with Unreliable Teachers, CoRR. abs/2106.04928 (2021). https://arxiv.org/abs/2106.04928.

[54] N. Frosst, S. Sabour, G.E. Hinton, DARCCC: Detecting Adversaries by Reconstruction from Class Conditional Capsules, CoRR. abs/1811.06969 (2018). http://arxiv.org/abs/1811.06969.

[55] Y. Qin, N. Frosst, C. Raffel, G.W. Cottrell, G.E. Hinton, Deflecting Adversarial Attacks, CoRR. abs/2002.07405 (2020). https://arxiv.org/abs/2002.07405.

[56] P. Deng, M.S. Rahman, M. Wright, Detecting Adversarial Patches with Class Conditional Reconstruction Networks, CoRR. abs/2011.05850 (2020). https://arxiv.org/abs/2011.05850.

[57] D.D. Thang, T. Matsui, Adversarial Examples Identification in an End-to-End System With Image Transformation and Filters, IEEE Access. 8 (2020) 44426–44442. https://doi.org/10.1109/ACCESS.2020.2978056.

[58] Y. Su, G. Sun, W. Fan, X. Lu, Z. Liu, Cleaning Adversarial Perturbations via Residual Generative Network for Face Verification, in: IEEE Int. Conf. Acoust. Speech Signal Process. ICASSP 2019 Brighton U. K. May 12-17 2019, IEEE, 2019: pp. 2597–2601. https://doi.org/10.1109/ICASSP.2019.8683327.

[59] S. Gu, P. Yi, T. Zhu, Y. Yao, W. Wang, Detecting Adversarial Examples in Deep Neural Networks using Normalizing Filters, in: A.P. Rocha, L. Steels, H.J. van den Herik (Eds.), Proc. 11th Int. Conf. Agents Artif. Intell. ICAART 2019 Vol. 2 Prague Czech Repub. Febr. 19-21 2019, SciTePress, 2019: pp. 164–173. https://doi.org/10.5220/0007370301640173.

[60] W. Fan, G. Sun, Y. Su, Z. Liu, X. Lu, Hybrid Defense for Deep Neural Networks: An Integration of Detecting and Cleaning Adversarial Perturbations, in: IEEE Int. Conf. Multimed. Expo Workshop ICME Workshop 2019 Shanghai China July 8-12 2019, IEEE, 2019: pp. 210–215. https://doi.org/10.1109/ICMEW.2019.00-85.

[61] Y. Ruan, J. Dai, TwinNet: A Double Sub-Network Framework for Detecting Universal Adversarial Perturbations, Future Internet. 10 (2018) 26. https://doi.org/10.3390/fi10030026.

[62] M. AprilPyone, Y. Kinoshita, H. Kiya, Filtering Adversarial Noise with Double Quantization, in: 2019 Asia-Pac. Signal Inf. Process. Assoc. Annu. Summit Conf. APSIPA ASC 2019 Lanzhou China Novemb. 18-21 2019, IEEE, 2019: pp. 1745–1749. https://doi.org/10.1109/APSIPAASC47483.2019.9023341.

[63] H.W. Kuhn, The Hungarian method for the assignment problem, Nav. Res. Logist. Q. 2 (1955) 83–97. https://doi.org/10.1002/nav.3800020109.

[64] G.E. Hinton, O. Vinyals, J. Dean, Distilling the Knowledge in a Neural Network, CoRR. abs/1503.02531 (2015). http://arxiv.org/abs/1503.02531.

[65] A. Kosiorek, Stacked Capsule Autoencoders, (2020). https://github.com/akosiorek/stacked_capsule_autoencoders.

# Appendix

## A. Generating Adversarial Samples for Training

The generation algorithm of adversarial samples for training is based on the evasion attack algorithm proposed in section 4. We replace the optimizer in the original algorithm with the following formula to accelerate the generation of adversarial samples:

$$p'_{N+1} = p'_N - \beta \cdot \text{sign}\left(\nabla_{p'_N}\left(\|p_N\|_2 + \alpha \cdot f(x + p_N)\right)\right) \tag{9}$$

where $p_N$ represents the perturbation obtained in the $N^{th}$ iteration, $f(x + p_N)$ is the target function of Formula (3), and $\beta > 0$ is a hyperparameter which defines the step of update in each iteration. The whole algorithm is shown in Algorithm 5:

---

**Algorithm 5:** Generating Adversarial Samples for Training

1: **Input:** Image $x$, SCAE model $E$, classifier $C$, hyperparameter $\alpha$, hyperparameter $\beta$, the number of outer iterations $n_{o\_iter}$, the number of inner iterations $n_{i\_iter}$.
2: **Output:** Adversarial sample $x^*$.
3:
4: Initialize $\mathcal{L}^* \leftarrow +\infty$.
5: $S \leftarrow \left\{i \big| E(x)_i > \frac{1}{K}\sum_{j=1}^{K} E(x)_j\right\}$
6: $w \leftarrow \text{arctanh}(2x - 1)$
7: **for** $i$ in $n_{o\_iter}$ **do**
8:      $p'_0 \leftarrow \text{rand}(0,1)$
9:      **for** $j$ in $n_{i\_iter}$ **do**
10:          $x_j^{adv} \leftarrow \frac{1}{2}\left(\tanh(w + p'_j) + 1\right)$
11:          $\mathcal{L} \leftarrow \|x_j^{adv} - x\|_2 + \alpha \cdot \sum_{i \in S} E(x_j^{adv})_i$
12:          */* Computes next $p'_j$ */*
13:          $p'_{j+1} \leftarrow p'_j - \beta \cdot \text{sign}\left(\nabla_{p'_j}\mathcal{L}\right)$
14:          $x_{j+1}^{adv} \leftarrow \frac{1}{2}\left(\tanh(w + p'_{j+1}) + 1\right)$
15:          */* Judge if the current result is the best one */*
16:          **if** $C\left(E(x_{j+1}^{adv})\right) \neq C(E(x))$ **and** $\|x_{j+1}^{adv} - x\|_2 < \mathcal{L}^*$ **do**
17:              $\mathcal{L}^* \leftarrow \|x_{j+1}^{adv} - x\|_2$
18:              $x^* \leftarrow x_{j+1}^{adv}$
19:          **end if**
20:      **end for**

| | |
|---|---|
| 21: | Update $\alpha$. |
| 22: | **end for** |
| 23: | **return** $x^*$ |

The update strategy of $\alpha$ is the same as that described in section 4. We will not go into much detail here.